\documentclass[10pt,twocolumn,letterpaper]{article}
\usepackage{cvpr}             
\usepackage{graphicx}
\usepackage{amsmath}
\usepackage{amssymb}
\usepackage{booktabs}
\newcommand{\bfx}{\mathrm{x}}
\newcommand{\bfz}{\mathrm{z}}
\newcommand{\bfv}{\mathbf{v}}
\usepackage{times}
\usepackage{epsfig}
\usepackage{caption}
\usepackage{subcaption}
\usepackage{comment}
\usepackage{paralist}
\usepackage{enumitem}
\DeclareMathOperator*{\argminA}{arg\,min}
\usepackage{xcolor}

\usepackage[pagebackref,breaklinks,colorlinks]{hyperref}

\def\cvprPaperID{2466} 
\def\confName{CVPR}
\def\confYear{2022}

\title{NODEO: A Neural Ordinary Differential Equation Based Optimization Framework for Deformable Image Registration}

\author{
Yifan Wu\footnotemark
\quad Tom Z. Jiahao\footnotemark[\value{footnote}]
\quad Jiancong Wang
\quad Paul A. Yushkevich 
\quad M. Ani Hsieh 
\quad James C. Gee
\\
University of Pennsylvania, Philadelphia, PA, USA\\
{\tt\small \{yfwu, zjh\}@seas.upenn.edu}\quad
{\tt\small \{jiancong.wang, pauly2\}@pennmedicine.upenn.edu} \\
{\tt\small mya@seas.upenn.edu} \quad
{\tt\small gee@upenn.edu}
}

\begin{document}
\twocolumn[{%
\renewcommand\twocolumn[1][]{#1}%
\maketitle
\vspace{-2em}
\begin{center}\vspace{-1em}
    \centering
    \includegraphics[width=\textwidth]{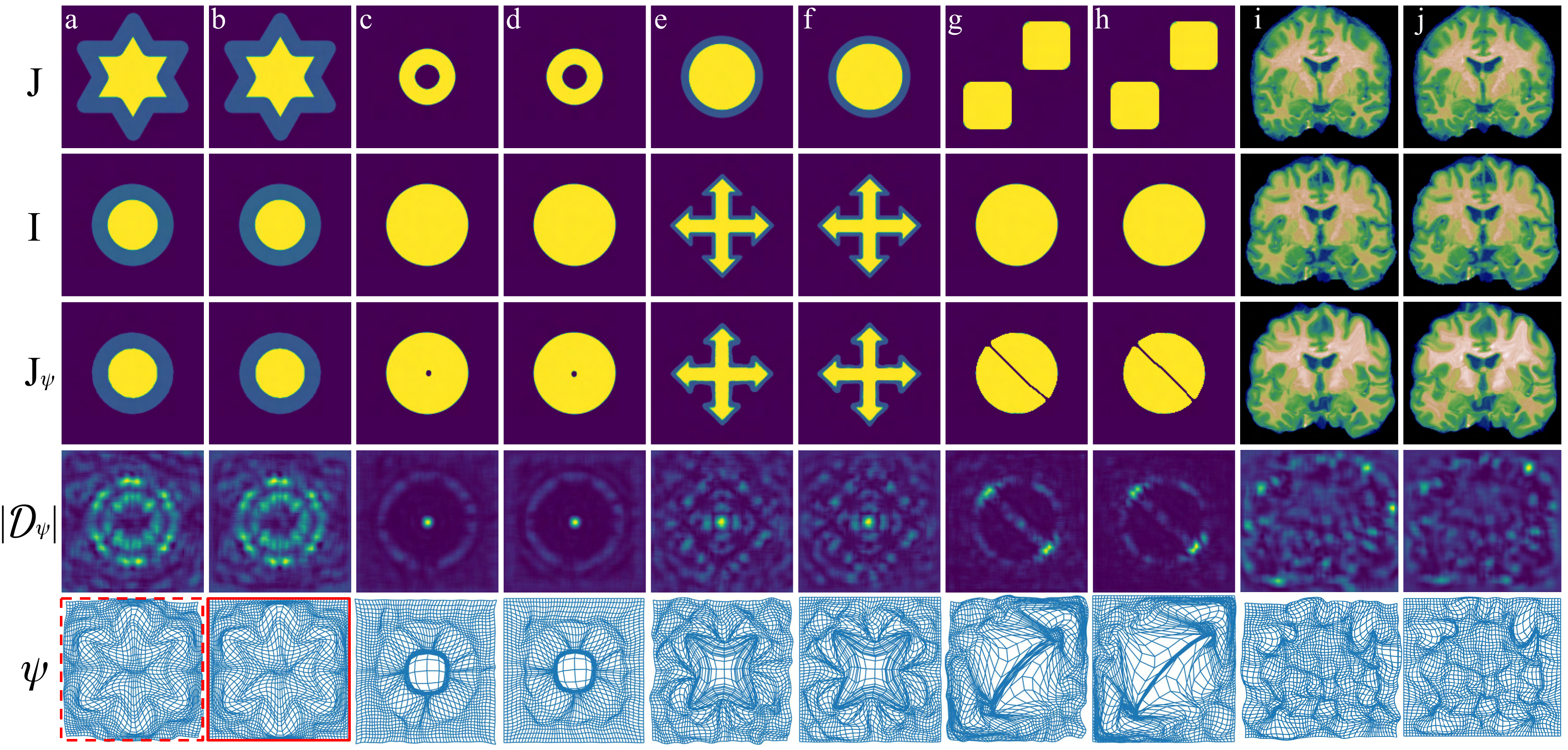}
    \captionof{figure}{\textbf{2D image registration using our framework.} The rows show ($J$) the moving images, ($I$) fixed images, ($J_\psi$) warped moving images, ($|\mathcal{D}_\psi|$) Jacobian determinants of the transformation $\psi$, and visualization of $\psi$ respectively. The columns (b)(d)(f)(h)(j) incorporate a fixed boundary constraint, as indicated by the box with solid red line, while (a)(c)(e)(g)(i) do not (box with dotted red line). }
    \label{fig: demo}
\end{center}
}]

\begin{abstract}
\footnotetext{\text{*}Equal contribution.}
\noindent Deformable image registration (DIR), aiming to find spatial correspondence between images, is one of the most critical problems in the domain of medical image analysis. In this paper, we present a novel, generic, and accurate diffeomorphic image registration framework that utilizes neural ordinary differential equations (NODEs). We model each voxel as a moving particle and consider the set of all voxels in a 3D image as a high-dimensional dynamical system whose trajectory determines the targeted deformation field. Our method leverages deep neural networks for their expressive power in modeling dynamical systems, and simultaneously optimizes for a dynamical system between the image pairs and the corresponding transformation. Our formulation allows various constraints to be imposed along the transformation to maintain desired regularities.  Our experiment results show that our method outperforms the benchmarks under various metrics. Additionally, we demonstrate the feasibility to expand our framework to register multiple image sets using a unified form of transformation, which could possibly serve a wider range of applications. Our project page: \url{https://yifannnwu.com/NODEO-DIR/}.
\end{abstract}
\section{Introduction}
\label{sec:intro}

\noindent
Deformable image registration (DIR) is a process for establishing spatial correspondences between images. The term ``deformable" points to the nonlinear and dense nature of the required transformation. DIR has a broad range of applications including normalization of population studies, quantifying changes in longitudinal imaging, accounting for motions of organ, as a building block of other image analysis algorithms, etc. 
While there is usually no ground-truth for the optimal transformation, 
image registration is usually formulated according to and evaluated on three criteria: the accuracy of matching the source and target images in terms of a pre-defined similarity metric, the regularity of the transformation to ensure that it is well-behaved, such as topology preservation, and the speed of the algorithm \cite{sotiras2013deformable, tustison2019learning}.

Traditional approaches solve DIR as a pair-wise optimization problem. These methods enforce transformation regularity through hard model assumptions. For example, Large Deformation Diffeomorphic Metric Mapping (LDDMM) \cite{beg2005computing, miller2006geodesic}, one of the most influential approaches, solves for diffeomorphisms by formulating the registration problem using a flow Partial Differential Equation (PDE), where the intergral of time-varying velocity fields produces the final deformation. 
These methods, however, can be challenging to apply if both speed and accuracy are desired, and they potentially limit the performance if different model assumptions are needed \cite{sotiras2013deformable}. 

With rapid advancement in machine learning and abundance of medical imaging data, there is increased interest in developing deep learning based methods to solve the DIR problem ~\cite{VoxelMorph_v1, VoxelMorph_v4, mok2020fast, dalca2019learning, yang2017quicksilver}. These methods significantly 
reduce the runtime of registration through using neural networks to learn a good sharing representation of a training dataset. Registration on a new image pair then becomes a rapid inference process. 
However, the generalizablity to unseen data is a long-standing challenge of recent data driven learning based methods, precluding their straightforward application in practice. 

Recent development in scientific machine learning has shown promising results in modeling differential equations using neural networks \cite{zhuang2021multiple, Jiahao2021KnowledgebasedLO}, which can describe any system that evolves with ``time". Given the demonstrated utility of flow-based approaches as in LDDMM\cite{beg2005computing} and the known advantages in expressive power of deep neural networks, we ask: \emph{can we integrate the merits of both?} We attempt to analyze this possibility from the following perspectives.

\textbf{(i)} The image registration problem can be viewed as a system-identification problem. Specifically, our task is to find a differential equation whose solution gives the transformation between the images. By imposing few assumptions on the dynamical system, we are able to explore various classes of systems and their solutions to achieve the desired registration while ensuring certain solution properties on demand, such as topology preservation.

\textbf{(ii)} Traditional model-based approaches are naturally constrained by the models they assume and implement. The hard constraints imposed by a given model may potentially be relaxed for better model capacity and regularity. In particular, a learnable flow that allows for penalties on its trajectory may serve as an alternative framework.

\textbf{(iii)} Adopting deep learning in DIR may not require additional data. The expressive power of deep learning stems from its compositional rule of functions. Much of the existing DL-based registration work builds on the foundations of feature learning, which learns the map or flow from supervised labels or a sufficient number of image pairs. Another easily overlooked utility of neural networks is they can serve as a 'parametric' backbone for general optimization problems. For the DIR problem, we can use a network to parameterize the system we aim to identify in (i-ii).

Motivated by the analysis above, we propose \emph{NODEO}, a neural ordinary differential equation based optimization framework that formulates the velocity field optimization as a neural network optimization. Specifically, we treat the set of all pixel/voxel locations in an image as one single evolving system, and parametrize the system's evolving dynamics with a deep neural network. The registration task therefore becomes finding a system whose trajectory's end point is the deformation field that minimizes the dissimilarity between images.

The benefits of our approach are twofold. \textit{First}, starting from the generalized flow field approach and then specializing to image registration, our framework provides enhanced flexibility. Our framework makes embedding proper dynamical constraints (such as spatial smoothness) straightforward, allowing flexibility in terms of defining task-specific ``goodness" for our transformations, and producing solutions with the desired properties. 
We can also easily incorporate boundary conditions on demand. By imposing loss penalties on the intermediate states of the trajectory, our model permits great flexibility in the type and number of assumptions one can impose on the solution. Thus, our framework is straightforward to extend from image pairs to multiple-image sets by adding intermediate supervision. 
 \textit{Second},  we explicitly model the image grid as one high-dimensional system, so that convolutional layers can be naturally leveraged to allow particles to spatially interact with each other within the system dynamics. The proposed solution brings the full expressive power of neural nets, as demonstrated in Figure~\ref{fig: demo}. Note that our approach falls under the family of pair-wise optimization approaches, \textit{i.e.} network parameters are optimized without any additional data. 
 

Overall, the contributions of this paper are: (1) generalization of the flow field approach to registration through high-dimensional dynamical system modeling, (2) unifying the optimization problem of discovering differential equations and their solutions (velocity fields) in one network \textit{without training data}, (3) enhanced flexibility and effectiveness of adding desired regularizations and constraints on solution transformations, and (4) demonstration that the proposed framework can be serve as an alternative approach to optimization-based registration tool that have \textit{proven utility} across numerous application domains, with state-of-the-art performance over a variety of evaluation metrics. 

\section{Related Work}

\subsection{Pair-wise Optimization-based Methods}
There are a number of prominent DIR techniques that have been evolved into widely used tools.
NiftyReg \cite{modat2010fast} is a representative work of parametric approaches \cite{rueckert1999nonrigid, shen2002hammer} for describing continuous interpolations with a finite set of parameters utilizing basis functions, b-splines, etc. Demons and its variants \cite{thirion1996non, vercauteren2008symmetric, vercauteren2009diffeomorphic} are optical flow based non-parametric techniques. SyN \cite{avants2008symmetric} is a representative work of greedy techniques and is well-known for producing symmetric solutions. The solution transformations family is a choice one makes based on the application or features one seeks to develop. When it's desired that the tranformation can be large deformation and be topology preserving, which is a property useful in many applications, then flow formulation is the natural family of transformation to use. Large Deformation Diffeomorphic Metric Mapping (LDDMM) is one of the most influential method in this family, which generate the trajectory from the source to target images rather than simply the transformation. LDDMM formulates the registration problem as solving a velocity field that ``flows" the source image to the target one. Early LDDMM works directly solve the Euler-Lagrange equation by variational approach. Shooting method is adopted later using Euler-Poincare characteristic for the velocity field to guarantee geodesic path, reducing the optimization space from spatial-temporal velocity fields to initial momentum \cite{beg2005computing, miller2006geodesic}.  The elegant formulation of the LDDMM motivates a number of follow-up efforts to improve the framework, including developments of optimization method using adjoint \cite{vialard2012diffeomorphic}, discretize-then-optimize paradigm to reduce runtime \cite{polzin2020discretize, joshi2021diffeomorphic}, spatial-temporal variant regularizations to relax the constraints \cite{shen2019RDMM} and so on. LDDMM is tailored to the registration problem and makes strong assumptions about the dynamics of the flow field, whereas our method is more flexible because we begin with a generic flow field approach and then narrow down to image registration.

\subsection{Data-driven Learning-based Methods}
With the power of deep learning, there is growing interest in creating learning-based solutions to the DIR problem \cite{yang2017quicksilver, VoxelMorph_v1, VoxelMorph_v2, VoxelMorph_v4}. By learning a common representation for a collection of images then performing registration in the inference stage, learning-based methods can significantly reduce the runtime \cite{dey2021generative, dalca2019learning, mok2021conditional, shen2019networks}. Some efforts attempt to include regularity, such as diffeomorphism, into networks by developing symmetric and reversevable structures or penalty \cite{mok2020fast, greer2021icon, kim2021cyclemorph}. To better solve large deformations, Xu \etal proposed to use neural ODE on image registration to refine the estimated transformation, by modeling the dynamics of the parameters of registration models (e.g., b-splines) \cite{xu2021multi}. However, they do not use neural ODE to directly describe the transformation between images as a flow like ours. Despite numerous efforts to improve generalizability, such as data augmentation or low-shot learning \cite{shen2020anatomical, cui2020unified}, generalizability remains a long-standing difficulty of current data-driven learning-based approaches. In contrast to the most learning-based methods for registration, which build on feature learning, we employ network to represent a differential equation.

\section{Method}
\begin{figure*}[t]	
\small
	\begin{center}
		\includegraphics[width=\linewidth,keepaspectratio]{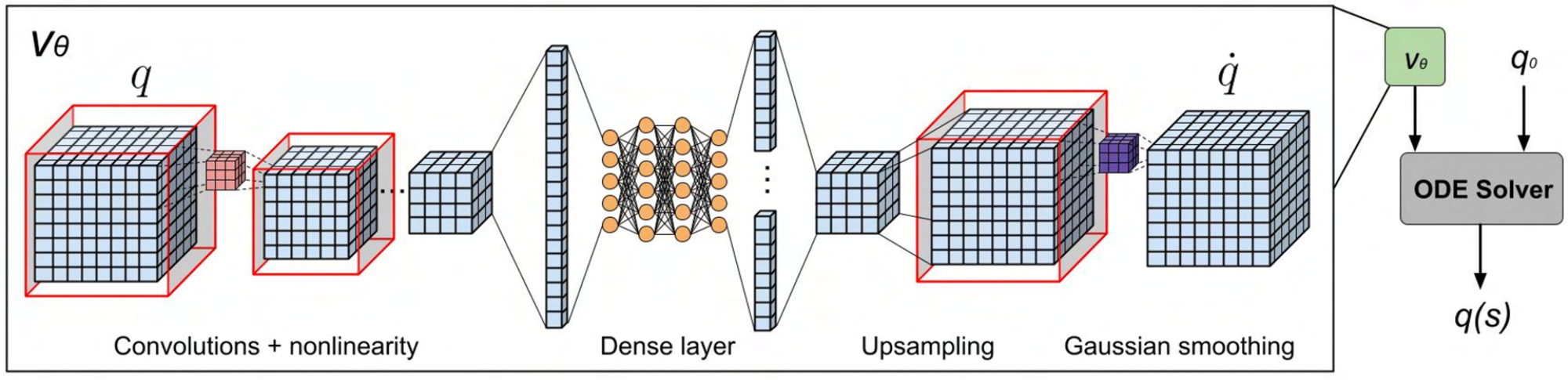}
	\end{center}
	\caption{
		\textbf{Framework overview.} Our frame models the vector field $\bfv_\theta$ as a neural network. A voxel cloud \textbf{without intensity} first gets down-sampled using convolutional layers. It then passes through dense layers where time is injected. It then gets upsampled and restored back to the shape of the voxel cloud and smoothened with a gaussian kernel. Note the images are only used for similarity measurement. }
	\label{fig:overview}
\end{figure*}

\subsection{Deformable Image Registration Formulation}
The deformable image registration problem can be formulated as follows: consider an unparametrized 3D image as a discrete solid, where the location of the $i^{th}$ voxel/point is given by $\bfx_i \in \Omega \subseteq \mathbb{R}^3, i\neq j \iff \bfx_i \neq \bfx_j$, where $\Omega$ is the image domain. This voxel location is known in shape analysis as a landmark coordinate. The location of all voxels or the voxel cloud can be denoted by the ordered set $q = \{x_i\}_{i=1}^{N}$, where $N =D \times H \times W$ is the total number of voxels in the image, and $ D, H$, and $W$ are the image depth, height and width respectively. As a shorthand, we write $\Pi$ as the domain of voxel clouds and $q\in\Pi$. We denote the fixed image by $I$ and the moving image by $J$, which are functions $I, J: \Omega \rightarrow \mathbb{R}^d$, mapping each voxel coordinate to the voxel value/intensity. In this work, we only consider scalar-valued, e.g., MRI images, and therefore $d=1$.

Traditionally, the goal of DIR is to find some \textit{good} transformation $\phi: \Omega \rightarrow \Omega$ such that the transformed moving image $J(\phi (\bfx)), \forall\bfx\in q$ is similar to the fixed image $I$~\cite{beg2005computing}. Here, $\phi$ is the spatial transformation that maps the domain of voxels $\Omega$ onto itself. An identity map is defined as $\phi_0$ such that $\phi_0(\bfx) = \bfx$. In many applications \cite{gee1993elastically} of DIR, a \textit{good} property of transformation $\phi$ is both sufficiently smooth and diffeomorphic in $\mathbb{R}^3$. The latter condition requires that the topology of the moving image is preserved under transformation. In other words, the transformation $\phi$ should not create 
folds in $\Omega$. We can overload the function $\phi$ by defining its application on a voxel cloud as a point-wise transformation on the voxels, i.e. $\phi(q_f) = (\phi(\bfx_1), \phi(\bfx_2), ..., \phi(\bfx_N))^T$, where $q_f = (\bfx_1, \bfx_2, ..., \bfx_N)^T$ is the spatially flattened voxel cloud. Similarly, this point-wise transformation can be defined for image $J$ when applied on a transformed voxel cloud. 

Following the above definitions, DIR is generally formulated as the minimization of the combined image similarity metric and regularization of the transformation given by

\begin{equation}
    \mathcal{J}(\phi; I, J) = \mathcal{S}(J(\phi(q_0)), I) + \mathcal{R}(\phi),
    \label{eqn: DIR traditional}
\end{equation}
where $q_0$ is the initial voxel cloud when none of the voxels in the cloud has undergone transformation. The term $\mathcal{S}(\cdot,\cdot)$ is the similarity metric that measures the difference between the deformed moving image $J$ and the fixed image $I$. The term $\mathcal{R}(\cdot)$ is the regularization on the transformation.

However, in this work, we reformulate the problem by searching for a transformation $\psi: \Pi \rightarrow \Pi$ which maps the domain of the voxel cloud $\Pi$ onto itself, while requiring diffeomorphism in $\Pi$. This is a marked contrast with 
most works on pointcloud deformation without intensity \cite{jiang2020shapeflow, niemeyer2019occupancy, yang2019pointflow} that model mapping $\mathbb{R}^3 \rightarrow \mathbb{R}^3$. In this work, we explicitly allow interactions among voxels by searching for a transformation in the high-dimensional cloud space. Our problem formulation has a similar form as Eqn. \eqref{eqn: DIR traditional}, with the exception that the transformation over the voxel cloud is no longer point-wise, but instead given by

\begin{equation}
    \mathcal{J}(\psi; I, J) = \mathcal{S}(J(\psi(q_0)), I) + \mathcal{R}(\psi),
    \label{eqn: DIR ours}
\end{equation}
where $\psi(q_f) = (\psi(\bfx_1, \bfx_2, ..., \bfx_N))^T$. We can alternatively write it as $\psi(q_f) = (\psi_1(q_f), \psi_2(q_f), ..., \psi_N(q_f))^T$, where $\psi_i$ is called the $i^{th}$ component of the $\psi$. 


\subsection{Neural Ordinary Differential Equations}
Taking inspiration from the resemblance between residual networks and dynamical systems, Chen et al.~\cite{chen2018neural} first introduced neural ordinary differential equations (NODEs) to approximate infinite depth neural networks. It aims to learn the function $f$ parameterized by $\theta$ by defining a loss function of the following form
\begin{equation}
\begin{aligned}
    \frac{dz}{dt} = f_{\theta} (z(t), t),
    \label{eqn: NODE1}
\end{aligned}
\end{equation}

\begin{equation}
\begin{aligned}
    \mathcal{L}(\bfz(t_1)) = \mathcal{L}\left(\bfz_0 + \int^{t_1}_{t_0} f_{\theta}(\bfz(t), t)dt\right).
    \label{eqn: NODE2}
\end{aligned}
\end{equation}

From a system perspective, NODEs are continuous-time models that represent vector fields as neural networks. It has since been adapted as a universal framework for modeling high-dimensional spatiotemporally chaotic systems utilizing convolutional layers~\cite{Jiahao2021KnowledgebasedLO}, demonstrating its ability to capture highly complex behaviors in space and time. Hence, we find it a suitable candidate for our registration task. 

Since NODEs often require the numerical solver to take many steps to realize the flows, they are memory-inefficient if all gradients along the the integration steps needs to be stored using traditional backpropagation. Hence many recent works~\cite{chen2018neural, zhuang2020adaptive} on NODEs have therefore focused on reducing the memory requirements for gradient propagation. Notably, the adjoint sensitivity method (ASM) has enabled constant memory gradient propagation for optimizing NODEs, and we adopt ASM in our work as well. For a brief description of ASM, one can refer to the supplementary. Proofs for its gradient convergence can be found in~\cite{chen2018neural, Jiahao2021KnowledgebasedLO}. 

ASM enables our framework to interpolate between $t=0$ and $t=s$ for an arbitrary number of steps with constant memory cost. This is particularly helpful when a temporally smooth diffeomorphic flow is required, as the numerical solver can increase its number of steps to improve the smoothness of $q(t)$ with respect to $t$. Also, our model can also be extended to multiple or sequential images by imposing constraints on the intermediate states of the trajectory.

\subsection{DIR in Dynamical System View }
Our work borrows intuition from dynamical systems and treats the trajectory of the entire voxel cloud as the solution to a first-order non-autonomous ordinary differential equation given by
\begin{equation}
\begin{aligned}
    \frac{dq}{dt} = \mathcal{K} \bfv_{\theta} (q(t), t),\\ 
    s.t.\ q(0) = q_0,
    \label{eqn: psi diff}
\end{aligned}
\end{equation}
where $\bfv_{\theta}(\cdot)$, as parametrized by $\theta$, is the vector field describing the dynamics of the voxel cloud, 
$q_0$ is the initial condition at $t=0$. We employ Gaussian kernels (for $\Omega \subseteq \mathbb{R}^3$, we use 3D Gaussian kernels), denoted by $\mathcal{K}$, as a filtering operator to enforce spatial smoothness in $\Omega$. Intuitively $\mathcal{K}$ is to ensure that the velocities of voxels are similar to those of their neighbors. Increasing the kernel size amounts to smoothing over larger voxel space, and therefore will improve the smoothness of the resulting flow; increasing the variance of the kernel amounts to encouraging more individual movements and therefore reduces the smoothness of the resulting flow.

The term \textit{non-autonomous}, or equivalently \textit{time-variant}, \textit{non-stationary} means that the time derivative of $q$ explicitly depends on $t$~\cite{strogatz2018nonlinear}. In other words, the velocity field attached to the Eulerian frame changes with time. 

The trajectory of $q$ is generated by integrating the ODE in Eqn. \eqref{eqn: psi diff} with the initial condition $q_0$. Assuming that the voxel cloud evolves from $t=0$ to $t=s$, the resulting voxel cloud at $t=s$ denotes the transformation $\psi(q_0)$ given by

\begin{equation}
    \psi(q_0) = q(s) = q_0 + \int^{s}_{0} \mathcal{K} \bfv_{\theta}(q(t), t) dt.
    \label{eqn: int}
\end{equation}
Eqn. \eqref{eqn: int} is referred to as a diffeomorphic flow map in dynamical systems. While the uniqueness and existence theorem~\cite{strogatz2018nonlinear} only implies diffeomorphism in the high-dimensional space $\Pi$, we will show that diffeomorphism can be achieved in $\Omega$ by incorporating soft constraints in the optimization task. In practice, the flow map is computed using a numerical integration scheme such as the Euler's method. Note that while $s$ is chosen to be $1$ in most existing works, it can be parametrized by the total number of steps taken by the solver and the corresponding step sizes. The task of finding the transformation $\psi$ therefore becomes finding the best set of parameters describing $\bfv$. The optimization problem therefore becomes:

\begin{equation}
\begin{aligned}
\small
    \theta = \argminA_{\theta \in \Theta} \mathcal{L}_{sim}\left(I, J(q_0 + \int^{s}_{0} \mathcal{K} \bfv_{\theta}\left(q(t), t\right) dt)\right)\\
    + \mathcal{R}(\psi, \bfv_{\theta})  + \mathcal{B}(\psi),
    \label{eqn: our optim}
\end{aligned}
\end{equation}
where $\Theta$ is the space of all possible parameters. The different components in the loss function include the similarity metric $\mathcal{L}_{sim}$, the regularizers $\mathcal{R}$, and the boundary conditions $\mathcal{B}$. The individual tasks can employ a different measure that suits the problem for each regularization term here.
The similarity loss is $\mathcal{L}_{sim}(I, J) = 1 - NCC(I, J)$, where $NCC$ is the normalized cross correlation given by

\begin{equation}
\begin{array}{l}
{NCC}(I, J)=\\
\frac{1}{N} \sum_{\bfx \in q(s)} \frac{\sum_{\bfx_{i}\in W}(I(\bfx_{i})-\bar{I}(\bfx))(J(\bfx_{i})-\bar{J}(\bfx))}{\sqrt{\sum_{\bfx_{i}\in W}(I(\bfx_{i})-\bar{I}(\bfx))^{2} \sum_{\bfx_{i}\in W}(J(\bfx_{i})-\bar{J}(\bfx))^{2}}},
\end{array}
\label{eqn: NCC}
\end{equation}
where $\bar{I}(\bfx)$ and $\bar{J}(\bfx)$ are the local mean of a size $w^3$ window $W$ with $\bfx$ being at its center position, and $\bfx_{i}$ is an element within this window. In this work we set $w$ as 21.

The regularization term consists of three terms: 
\begin{equation}
\label{eqn: R}
    \mathcal{R}(\psi, \bfv_\theta) = \lambda_1\mathcal{L}_{Jdet} + \lambda_2\mathcal{L}_{mag} + \lambda_3\mathcal{L}_{smt}.
\end{equation}
The first term, $\mathcal{L}_{Jdet}$, penalizes negative Jacobian determinants in the transformed voxel cloud and is given by 

\begin{equation}
\mathcal{L}_{Jdet}=\frac{1}{N} \sum_{\bfx \in q(s)}  \|\sigma(-(|\mathcal{D}_{\psi}(\bfx)|+\epsilon))\|_{2}^{2},
\label{eqn: Jdet}
\end{equation}
where $\mathcal{D}_{\psi}(\bfx)$ is the Jacobian matrix at $\bfx$ under the transformation $\psi$. Here $\sigma(\cdot)=max(0, \cdot)$ is the ReLU activation function, which is used to select only negative Jacobian determinants. If there are no 
folds in the transformation, its jacobian determinant $\mathcal{D}_{\psi}(\bfx)$ should be positive. Lastly, we add a small number $\epsilon$ to the Jacobian determinants as an overcorrection. 
Instead of using $L1$ regularization as in \cite{mok2020fast,VoxelMorph_v4}, we use $L2$ norm here. Regularization with $L1$ introduces sparsity, reducing the number of folds, while the $L2$ norm can minimize the overall magnitude of folds, thereby avoiding outliers. To adapt to a specific task, one can combine the two.
In our framework, $\mathcal{L}_{Jdet}$ is a critical component since it ensures that the flow is diffeomorphic in $\Omega$. In this work, the Jacobian matrix is implemented using the finite difference approximation.

\begin{figure*}[t]	
\small
	\begin{center}
		\includegraphics[width=\linewidth]{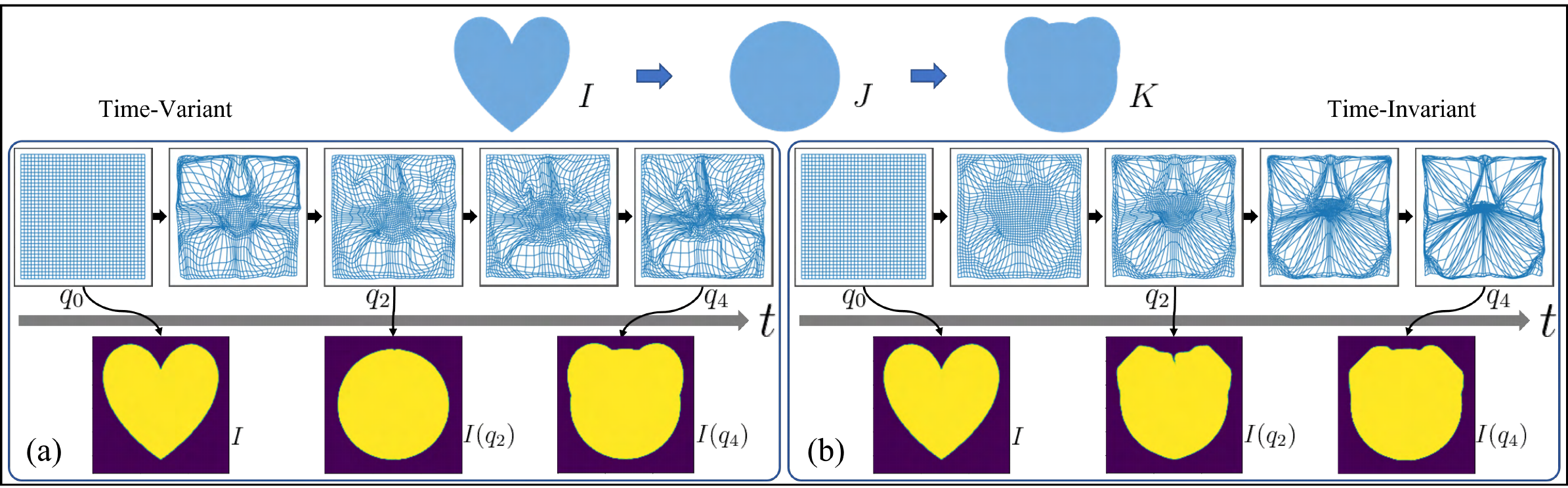}
	\end{center}
	\caption{
		\textbf{Discovering transformation on multiple images.}  The target images $I, J, K$ are shown in the top row, and the model is tasked to identify a path of transformation from $I$ to $K$ via $J$. The pictured registration results are from using (a) a time-varying system with explicit time embedding, and (b) a time-invariant system where the transformation does not depend on time. The transformations $\psi(q_t)$ are plotted on top of the warped images during the registration.}
		\vspace{-4mm}
    \label{fig: multiple}
\end{figure*}

The second term, $\mathcal{L}_{mag}$, regularizes the magnitude of the velocity field along the voxel cloud trajectory and is given by 
\begin{equation}
\mathcal{L}_{mag}= \frac{1}{N}\int^s_0 \|\mathcal{K}\bfv_{\theta}(q(t), t)\|_{2}^{2}dt.
\label{eqn: vmag}
\end{equation}
This amounts to penalising the \textit{energy} of the flow. In practice, the integral is replaced by a summation of the squared norm along the steps taken by the numerical integration scheme.  Lastly, the third term, $\mathcal{L}_{smt}$, is used to regularize the spatial gradients of the transformed voxel cloud and is given by 
\begin{equation}
\mathcal{L}_{smt}= \frac{1}{N} \sum_{\bfx \in q(s)}(\|\nabla_{\psi}(\bfx)\|_{2}^{2}),
\label{eqn: smt}
\end{equation}
where $\nabla_\psi(\bfx)$ denotes the spatial gradient around $\bfx$ under the transformation $\psi$. This encourages spatial smoothness of the transformed voxel cloud.  Similar to $\mathcal{L}_{Jdet}$, $\mathcal{L}_{smt}$ is implemented as a discrete approximation. Note that, even though $\bfv_\theta$ already includes Gaussian filtering, which in turn translates to the spatial smoothness of $\psi$, the inclusion of $\mathcal{L}_{smt}$ can reduce the degradation in smoothness as a result of numerical integration.
The last term in Eqn.\eqref{eqn: our optim}, $\mathcal{B}(\psi)$, specifies the boundary condition for the transformation $\psi$. While our MRI registration tasks do not specify any boundary condition ($\mathcal{B}(\psi) = 0$), we will demonstrate its effect through illustrative experiments on 2D images.


\section{Method Analysis}
\subsection{Illustrative Examples in 2D pair images}

To demonstrate the properties and capabilities of our framework, we used it to perform registration on a variety of 2D images ($\Omega \subseteq \mathbb{R}^2$) as shown in Figure \ref{fig: demo}. The 2D brain images have size $144\times 160$, and are slices taken from real brain MRI. All other images have size $144\times 144$, and were hand-drawn. We used Mean Squared Error (MSE) for similarity measurement in Eqn~\eqref{eqn: our optim}, and did not include $\mathcal{L}_{mag}$ and $\mathcal{L}_{smt}$. Based on Figure \ref{fig: demo}, we note that the resulting transformation $\psi$ is

\noindent\textbf{topology preserving:} While the registration warps the moving image as much as possible to look like the fixed image, it preserves the topology of the moving image in 2D. Columns (c) and (d) show that the warped ``donut'' closely resemble the circle, but the hole in its middle remains. Columns (g) and (h) also illustrate the 
two disconnected components will not become connected.

\noindent\textbf{diffeomorphic in 2D:} Diffeomorphism is a stronger condition than the topology preservation since it requires the transformation to be both continuous and differentiable. There are no visible 
folds in the visualization of $\psi$. There are few violations if we inspect the Jacobian determinants closely, and this can be further reduced by increasing the weight on $\mathcal{L}_{Jdet}$. Note that there actually exists no diffeomorphism between $I$ and $J$, for the examples in columns (e)(f), because the sharp corners make $I$'s manifold non-smooth. Even so, our framework produces a slightly rounded ``cross", as the result of a smooth approximation.

\noindent\textbf{enforcing boundary conditions:} Columns (a) (c) (e) (g) (i) show registration without boundary conditions, while the remaining columns fix the grids on all four sides. It can be observed that with this boundary condition, the four sides of the resulting $\psi$ remain unchanged. 

\begin{figure}[t]	
\small
	\begin{center}
		\includegraphics[width=0.95\linewidth,keepaspectratio]{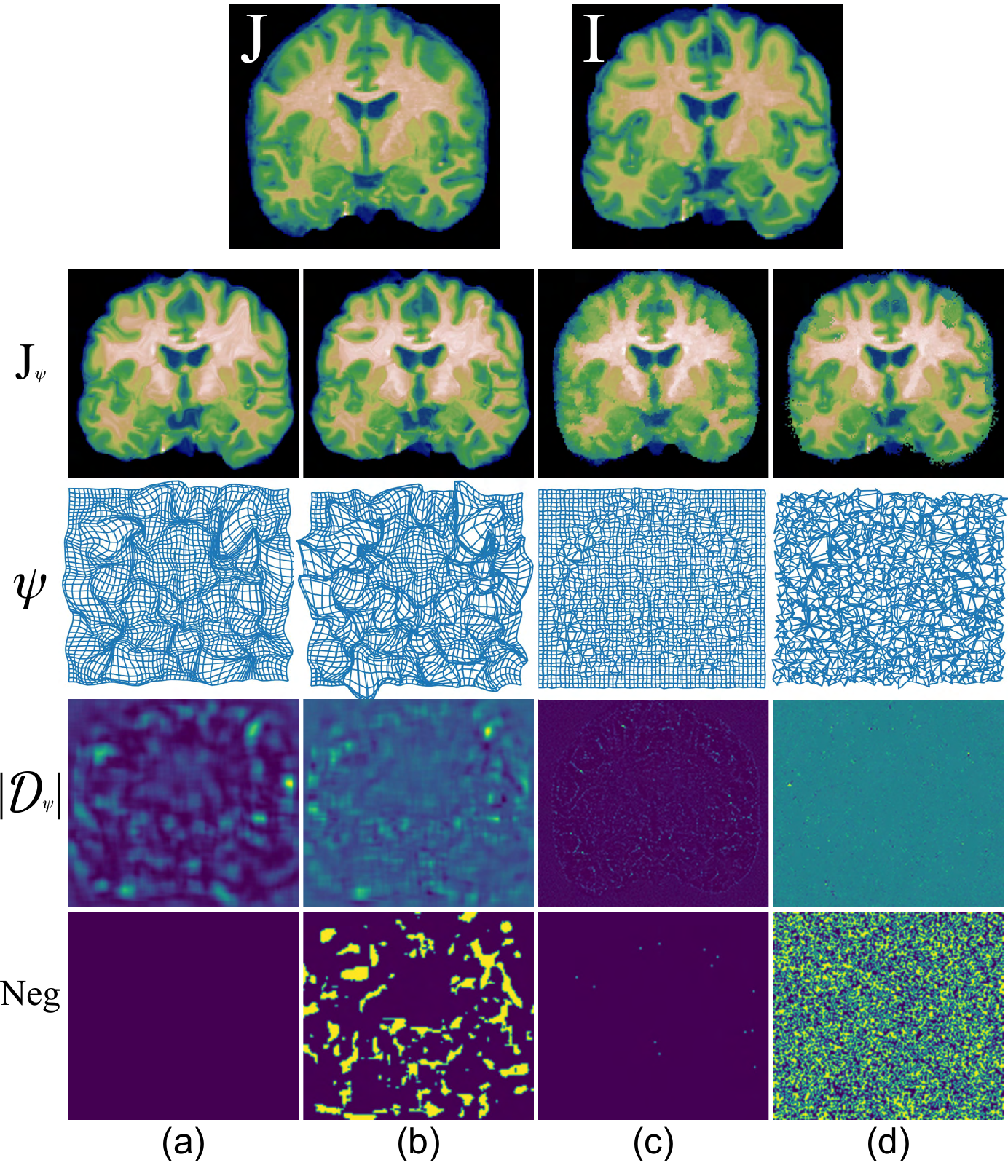}
	\end{center}
	\vspace{-2em}
	\caption{Demonstration of the effect of Gaussian smoothing $\mathcal{K}$ and the $\mathcal{L}_{Jdet}$ regularizer. The rows show ($J_\psi$) warped moving images, ($\psi$) grid visualization of the deformation field, ($|\mathcal{D}_\psi|$) Jacobian determinants of the transformation $\psi$, and (Neg) the regions with negative Jacobian determinants (yellow). The columns shows registration with (a) both $\mathcal{K}$ and $\mathcal{L}_{Jdet}$, (b) only $\mathcal{K}$, (c) only $\mathcal{L}_{Jdet}$, and (d) with neither.} 
	\label{fig:ablation2D}
\end{figure}

To analyze the effects of regularization terms, we conducted ablation studies. Figure \ref{fig:ablation2D} shows the effect of Gaussian smoothing and $\mathcal{L}_{Jdet}$ regularizer on 2D images. 
We discovered by applying hard constraints $\mathcal{K}$ to the model allows it to attain the requisite spatial smoothness and continuity. Even though we are modeling a high-dimensional system to take advantage of its expressive capacity, we can achieve diffeomorphic transformation in low dimension, $\mathbb{R}^2$ in this case, by using soft regularizer $\mathcal{L}_{Jdet}$.

\subsection{Illustrative Examples in 2D for multiple images}
Our framework can also be extended to perform registration on a sequence of images, where the intermediate images act as constraints along the transformations. Figure \ref{fig: multiple} shows the task of finding the transformation from a heart to a bear, and the image must take the shape of a circle during the intermediate step. Here, we compare a time-variant and time-invariant model. To make the system time-variant, we embed temporal information using positional encoding similar to that of a transformer model \cite{vaswani2017attention}. To ensure diffeomorphism during the entire transformation, we applied Eqn. \eqref{eqn: Jdet} on each of the intermediate steps. It can be seen from Figure \ref{fig: multiple} that the time-variant system produces a better registration result, demonstrating that incorporating time is important for more constrained tasks such as this. In practice, these intermediate images can be known temporal dynamics between image pairs (\textit{e.g.} infant development \cite{wang2019developmental}, disease progression \cite{adler2018characterizing}, cardiac or respiratory motions). In other words, being able to incorporate intermediate image constraints will allow for embedding domain knowledge into the registration process, and give more accurate transformations.

\section{Comparison with Benchmarks}
\subsection{Datasets and Pre-processing}
We evaluate our framework on the same registration task as the state-of-the-art \cite{mok2020fast}, which employs atlas-based registration. In this work, the atlases/templates are used as fixed images. We randomly select a small number of images as atlases and use the remainder as moving images to be registered to fixed images. We conduct experiments on two datasets: OASIS and CANDI.

\textbf{OASIS:} The OASIS \cite{marcus2007open} dataset consists of a collection of cross-sectional T1 MRI scans from 416 subjects. These subjects are aged 18-96 where 100 of them have been clinically diagnosed with mild to moderate Alzheimer’s disease. We use standard processing tool FreeSurfer \cite{fischl2012freesurfer} to resample all scans to $1mm\times 1mm \times 1mm$, followed by motion correction and skull stripping. We use the whole brain auto-segmentation provided in FreeSurfer for evaluation. Then we align all images 
to MNI 152 space \cite{fischl2012freesurfer} by affine transformation. The final images have a size of $160\times 192 \times 144$ by center cropping. We set five images with IDs $1, 10, 20, 30, 40$ as the atlases, and the remaining images with IDs $<50$, as moving images. The total number of moving images is $40$, resulting in $200$ pairs to be registered.

\textbf{CANDI:} The Child and Adolescent NeuroDevelopment Initiative (CANDI) \cite{kennedy2012candishare} dataset contains T1 MRI scans of 4 groups, which are Healthy Controls, Schizophrenia Spectrum, Bipolar Disorder, and Bipolar Disorder without Psychosis. We use manually-labeled whole brain segmentation provided in the dataset for skull stripping. Then we transform images along with their respective segmentations to the MNI 152 space and center crop the images to the size $160\times 192 \times 144$ similar to the OASIS set. We set the first subject of each group as the atlases (fixed images), and the following 5 subjects as moving images. The total number of moving images is $20$, resulting in $80$ pairs for registration.

\subsection{Evaluation Metric}
The goal of DIR is to find the spatial correspondence such that the \textit{similarity} of two images is maximized. In diffeomorphic registration, voxels are not allowed to self intersect, which can be guaranteed when the determinants of the Jacobian of the deformation field $\mathcal{D}_{\psi}(\bfx)$ are non-negative. We follow the convention \cite{VoxelMorph_v2, VoxelMorph_v4, mok2020fast, greer2021icon} and measure similarity and diffeomorphism using the following two criteria.
 
\textbf{Dice Similarity Coefficient :} The dice score measures the ratio between the overlap and union of two spatial regions. Here we calculate dice using the whole-brain segmentation maps. In particular, we evaluate dice between the fixed segmentation and the warped moving segmentation based on the deformation field given by the registration of the two images. We use auto-segmentation maps and compute the average dice across 28 anatomical structures as in VoxelMorph \cite{VoxelMorph_v1}. For the CANDI dataset, as it provides manual labels for 32 structures, we report the average dice for both 28 and 32 structures. One can refer to supplementary material for details of these anatomical structures.

\textbf{Jacobian Determinant:} In our experiment, we report the negative Jacobian ratio denoted as $r^\mathcal{D}$, which represents the number of voxels with negative Jacobian determinants versus the total number of voxels for each image. In the meanwhile, we report the sum of negative values of the Jacobian determinant, denoted as $s^\mathcal{D}$, which represents the total area/volume of folding in pixel/voxel unit. 

\subsection{Results}
We compare our proposed method with a state-of-the-art learning-based method SYMNet~\cite{mok2020fast}, and pair-wise optimization-based methods with well-developed software packages including SyN~\cite{avants2008symmetric}, Log-Demons~\cite{vercauteren2009diffeomorphic} and NiftyReg~\cite{modat2010fast}. SYMNet is considered the leading learning based framework for image registration since it outperforms other existing learning based methods such as the series of VoxelMorph works \cite{VoxelMorph_v1, VoxelMorph_v2}. Using a deep learning framework to learn the symmetric deformation fields, SYMNet achieves reversible registration and yields the best performance in terms of registration accuracy (dice) and quality of the deformation fields measured by $r^\mathcal{D}$ and $s^\mathcal{D}$. To enable a fair comparison with SYMNet, we employ the same dataset, {\it e.g.}, OASIS, and data processing practices and use the pre-trained model provided in the official SYMNet repository.

\begin{table}[t]
\caption{\textbf{Comparison with benchmarks.} The top part shows results on our OASIS data setting, dice is averaged over 28 structures. The bottom part shows results on our CANDI data setting, we report both mean dice on 28 and 32 structures. Numbers here are represented as mean or mean $\pm$ std. Note the result on OASIS of SYMNet is from the original paper \cite{mok2020fast}. }	
	\centering
	\resizebox{\linewidth}{!}{
		\begin{tabular}{c|ccc}
			\hline\hline
			\textbf{OASIS} dataset              &  Avg. Dice (28) $\uparrow$ &  $\mathcal{D}_{\psi}(\bfx)\leq 0$ $(r^\mathcal{D}) \downarrow$   & $\mathcal{D}_{\psi}(\bfx)\leq 0$ $(s^\mathcal{D}) \downarrow$  \\ \hline
			SYMNet \cite{mok2020fast}     & $0.743 \pm 0.113$ & $0.026\%$ & -\\             
			SyN \cite{ants_github}  & $0.729 \pm 0.109$   & $0.026\%$  &  $0.005$ \\
			NiftyReg \cite{niftyreg}  & $0.775 \pm 0.087 $ & $0.102\%$ & $1395.988$\\
			Log-Demons \cite{greedy} & $0.764 \pm 0.098 $ & $0.121\%$ & $84.904$  \\
			NODEO (ours $\lambda_1 = 2.5$) & $0.778 \pm 0.026$ &  $0.030\% $ & $34.183$ \\
			NODEO (ours $\lambda_1 = 2$) &\textbf{0.779 $\pm$ 0.026}& $0.030\% $ & 61.105 \\ \hline
			\textbf{CANDI} dataset	            & Avg. Dice (28) $\uparrow$ & $\mathcal{D}_{\psi}(\bfx)\leq 0 $ $(r^\mathcal{D}) \downarrow$  & $\mathcal{D}_{\psi}(\bfx)\leq 0$ $(s^\mathcal{D}) \downarrow$  \\ \hline
			SYMNet \cite{mok2020fast}  & $0.778 \pm 0.091$ & $1.4\times10^{-4}\% $ & $1.043$ \\  
			SyN  \cite{ants_github}  & $0.739 \pm 0.102$ & $ 0.018\% $ & $0.012$\\
			NiftyReg \cite{niftyreg}  & $0.775 \pm 0.088$ & $0.101\%$ & $1395.987$ \\
			Log-Demons \cite{greedy} & $ 0.786\pm 0.094 $ &  $0.071$ & $49.274$ \\
			NODEO (ours $\lambda_1 = 2.5$) & $0.801 \pm 0.011$ & $7.5\times10^{-8}\%$ & $1.574$\\
            NODEO (ours $\lambda_1 = 2$)    &\textbf{0.802 $\pm$ 0.011} & $1.8\times10^{-7}\% $ & $4.341$ \\ \hline
             \textbf{CANDI} dataset	            & Avg. Dice (32) $\uparrow$ & $\mathcal{D}_{\psi}(\bfx)\leq 0 $ $(r^\mathcal{D}) \downarrow$  & $\mathcal{D}_{\psi}(\bfx)\leq 0$ $(s^\mathcal{D}) \downarrow$  \\ \hline
            SYMNet \cite{mok2020fast}  & $0.736 \pm 0.015 $ & $1.4\times10^{-4}\% $ & $1.043$ \\  
			SyN \cite{ants_github}    & $0.713 \pm 0.177$ & $ 0.018\% $ & $0.012$ \\
			NiftyReg \cite{niftyreg}  & $0.748 \pm 0.160$ & $0.101\%$ & $1395.987$ \\
			Log-Demons \cite{greedy}  & $0.744 \pm 0.160$ & $0.071$ & $49.274$ \\
			NODEO (ours $\lambda_1 = 2.5$) &\textbf{0.760 $\pm$ 0.011}   & $7.5\times10^{-8}\%$ & $1.574$ \\
            NODEO (ours $\lambda_1 = 2$)   & \textbf{0.760 $\pm$ 0.011} & $1.8\times10^{-7}\% $ & $4.341$ \\
			\hline
		\end{tabular}
    }
	\label{table:SOTA}
\end{table}

\begin{figure}[t]	
\small
	\begin{center}
		\includegraphics[width=0.95\linewidth,keepaspectratio]{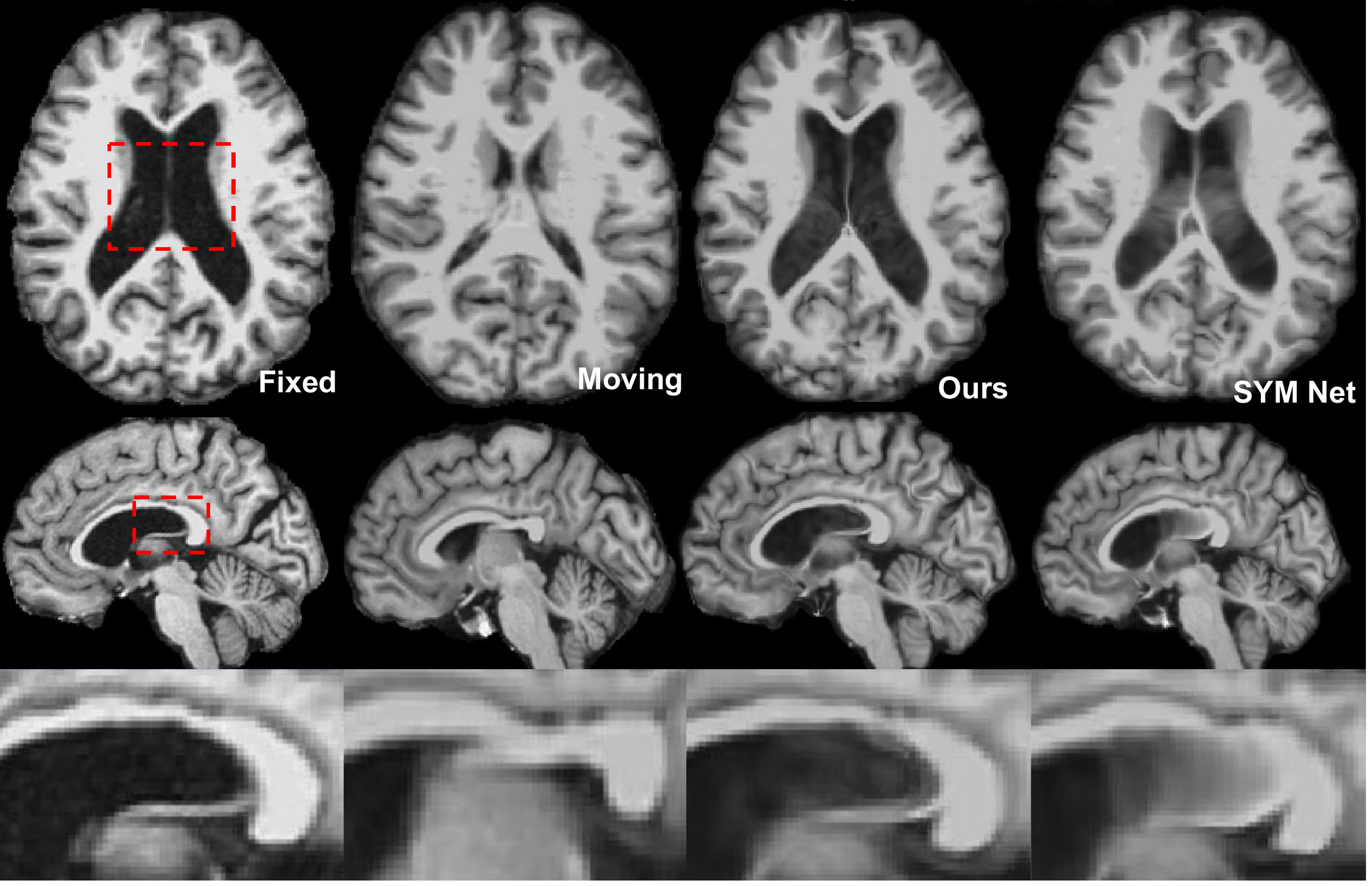}
	\end{center}
	\vspace{-2em}
	\caption{Images showing an example of a registration image pair. Fixed image is OASIS ID001 and Moving image is OASIS ID002. The 3rd column is ours warped image after registration and the 4th column is result of SYMNet \cite{mok2020fast}. 
	} 
	\label{fig:img_eg}
\end{figure}

The quantitative results are shown in Table~\ref{table:SOTA}. For both the OASIS and CANDI datasets, our method demonstrates a consistent and significant improvement in both mean dice scores over the 
brain structures as shown in Table~\ref{table:SOTA}. Our $r^\mathcal{D}$ (all $\leq 0.1 \%$) and low values of $s^\mathcal{D}$ verify that our method effectively produces diffeomorphic transformations. We visualize the qualitative result of the registration for one example pair image (OASIS-001 and OASIS-002) in Figure~\ref{fig:img_eg}. We can observe that the ventricular area of the warped brain image obtained with our method is much clearer and does not show any phantom artifacts (cloudy regions where it is black in the fixed image), which demonstrates that our method gives qualitatively better results. See the supplementary material for the full results of benchmarks. 

It is important to note that the OASIS dataset generally presents larger deformations between images compared with the CANDI dataset because of the larger age range of the subjects. Therefore we can see lower dice scores for the 28 structures in the OASIS set compared to the CANDI set and more folding in deformation fields. Also, note that 4 out of the 32 structures on CANDI are very small, which are inherently difficult for alignment in registration due to their lower spatial smoothness. This explains the lower dice on the 32 structures compared with that of the 28 structures. 

Lastly, we analyze the runtime and model complexity of our method and other benchmarks. Experiments shown in Table~\ref{table:SOTA} perform down-sampling on the image before passing it into the network in Figure~\ref{fig:overview}, reducing the runtime by $1/3$ without losing performance. The runtime for SYMNet, SyN, NiftyReg, Log-Demons and ours for one pair image registration are approximately 2s, 25 mins, 70s, 160s, and 80s. While the SYMNet (at inference) takes about 2 seconds to complete registration for one image pair, the pair-wise optimization-based methods show difficulty to achieve good performance and fast runtime simultaneously. In comparison, our method takes approximately 80 seconds to register a pair of image (1 step taken), and performs well in terms of similarity and regularity. 
The number of parameters in our model with the architecture illustrated in Figure~\ref{fig:overview} is $3/4$ of the number of voxels in the image, demonstrating that the expressive power is well presented without adding model complexity. For the experiments in Table~\ref{table:SOTA}, our method uses 3863 MB of memory on a 2080Ti GPU.

\vspace{-1em}
\section{Conclusion}
In this work, we proposed a generic framework for deformable image registration, and investigated the possibility of integrating the merits of both neural netowrks and flow formulations. The resulting models are flexible to incorporate desired transformation regularities and various constraints. We compared our method with benchmarks on several datasets and achieved state-of-the-art results under a variety of metrics. Future works include exploring different ways of time injection into the neural network and applying our methodology to sequential medical data.

\vspace{-1em}
\section{Acknowledgment}
This work was supported by NIH grants R01-NS096720, R01-HL133889, U24-MH114827, RF1-MH124605, RF1-AG069474, NSF IIS 1910308 and Office of Naval Research (ONR) Award No. 14-19-1-2253.

{\small
\bibliographystyle{ieee_fullname}
\bibliography{main}
}
\clearpage

\section{Supplementary Material}

\subsection{Adjoint sensitivity method (ASM) for neural ODE optimization}
Here we briefly introduce the Adjoint sensitivity method (ASM) for neural ODE optimization. While the loss function $\mathcal{L}$ in Eqn~\eqref{eqn: NODE2} can be any differentiable function, we will describe ASM by assuming $\mathcal{L}$ to be the mean squared error (MSE) between the resulting flow $\bfz(t_1)$ and the label $\bfz_l$ which is given by \eqref{eqn: NODE2}. The only reason we express MSE in this form is that it's more convenient for proving ASM convergence. We can therefore formulate the following optimization problem
\begin{equation}
\begin{aligned}
\min_{\theta} \quad & \mathcal{L}(\bfz(t_1)) = \int^{t_1}_{t_0}\delta(t_1-t)\|\bfz(t) - \bfz_l\|_2^2 dt, \\
\textrm{s.t.} \quad & \frac{d\bfz}{dt} = f_{\theta}(\bfz(t), t), \\
\quad & \bfz(t_0) = \bfz_0, 
\end{aligned}
\label{eqn: ASM optim}
\end{equation}
where $\delta(\cdot)$ is the Dirac delta function which ensures that only the gradients of the loss function with respect to $\bfz$ at $t = t_1$ gets propagated back. To propagate the gradient from the loss function to the parameters $\theta$, first we numerically solve the differential equation $d\bfz/dt = f_{\theta}(\bfz(t),t)$ for its trajectory forwards in time from $t_0$ to $t_1$ with the initial condition $\bfz(t_0) = \bfz_0$. Then we can define the adjoint equation given by:
\begin{equation}
    \frac{d\lambda^T}{dt} = - \mathbf{\lambda}^T \frac{\partial f}{\partial \bfz}\bigg\rvert_{\bfz = \bfz(t)} + \frac{\partial\mathcal{L}}{\partial\bfz}\bigg\rvert_{\bfz = \bfz(t)},
    \label{eqn: adjoint equation}
\end{equation}
where $\mathbf{\lambda}^T$ is a continuous-time Lagrange multiplier, also known as the adjoint variable. In the case of MSE, we have $\partial \mathcal{L}/\partial z = 2\cdot\delta(t_1-t)(\bfz(t) - z_l)$. We then numerically solve this equation backwards in time with the initial condition $\mathbf{\lambda}^T(t_1) = 0$ to obtain the trajectory of $\mathbf{\lambda}^T$ from $t=t_1$ to $t=t_0$. Lastly the gradient of the loss function with respect to the parameters, also known as the sensitivity is given by
\begin{equation}
    \frac{d\mathcal{L}}{d\theta} = -\int^{t_1}_{t_0} \mathbf{\lambda}^T \frac{\partial f}{\partial \theta}\bigg\rvert_{\theta = \theta(t)} dt.
    \label{eqn: NODEs gradient}
\end{equation}

After obtaining this gradient, we can then perform optimization with methods such as gradient descent. Note that the Jacobians $\partial f/\partial \bfx$ and $\partial f/\partial \theta$ can be computed efficiently using automatic differentiation during the forward pass. In summary, ASM solves for gradients through the following steps:
\begin{itemize}
    \item Numerically solve $d\bfz/dt = f_{\theta}(\bfz,t)$ forward in time from $t_0$ to $t_1$.
    \item Numerically solve the adjoint equation \eqref{eqn: adjoint equation} backward in time from $t_1$ to $t_0$ using the initial condition $\mathbf{\lambda}^T(t_1) = 0$.
    \item Numerically evaluate the integral in Eqn. \eqref{eqn: NODEs gradient} to obtain the desired gradient.
\end{itemize}

\subsection{Illustrative Examples in 2D pair images: results of LDDMM}
\begin{figure}[htb]	
    \small
	\begin{center}
		\includegraphics[width=\linewidth,keepaspectratio]{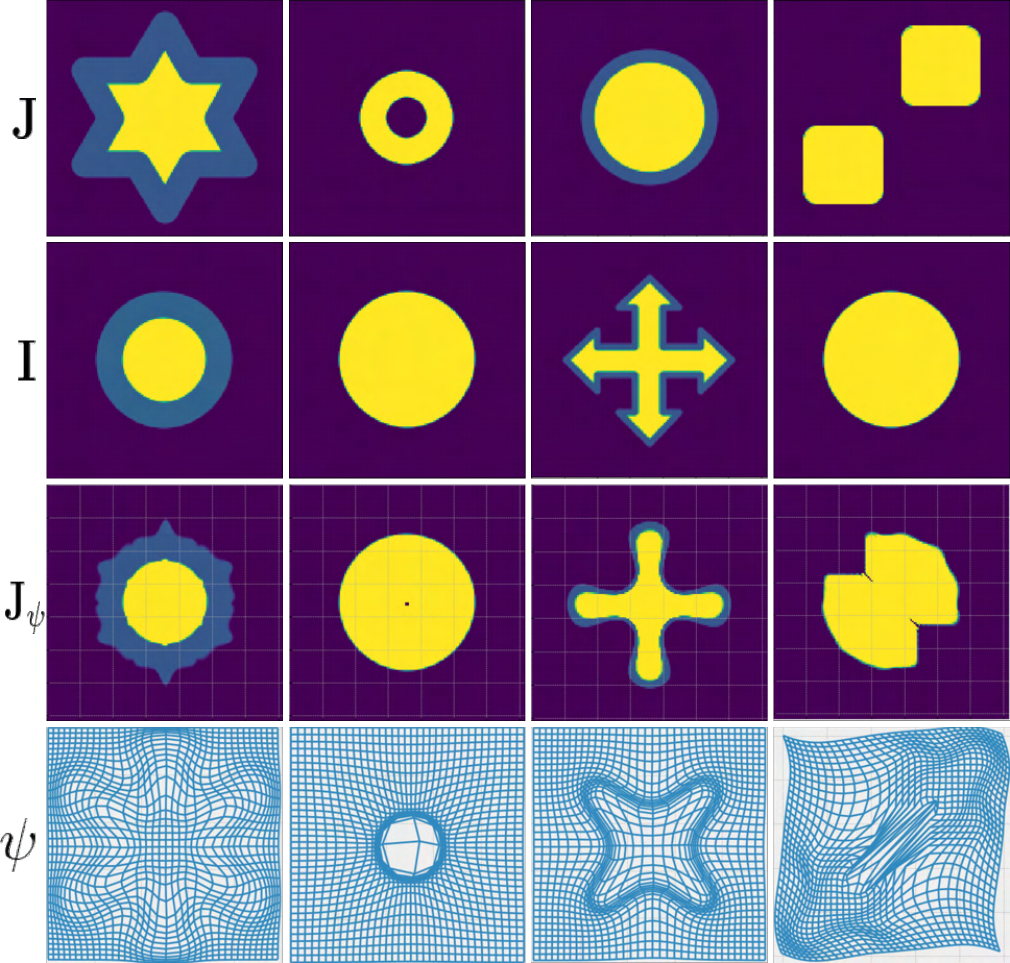}
	\end{center}
	\caption{
		The rows show ($J$) the moving images, ($I$) fixed images, ($J_\psi$) warped moving images, and visualization of $\psi$ respectively.}
	\label{fig:lddmm}
\end{figure}

To explore the different properties of solution transformations between LDDMM and ours, we conduct the same experiment on 2D pair examples as in Figure~\ref{fig: demo} using LDDMM shooting method. We use the Mermaid registration toolkit\footnote{https://github.com/uncbiag/mermaid}, with a learning rate of $0.01$ and 500 epoches of optimization. The results of LDDMM are shown in Figure~\ref{fig:lddmm}. LDDMM generates a smoother transformation field, while our method produces a more accurate match without violating diffeomorphism.

\subsection{Anatomical structures}
\begin{figure*}[htb]	
    \small
	\begin{center}
		\includegraphics[width=\linewidth,keepaspectratio]{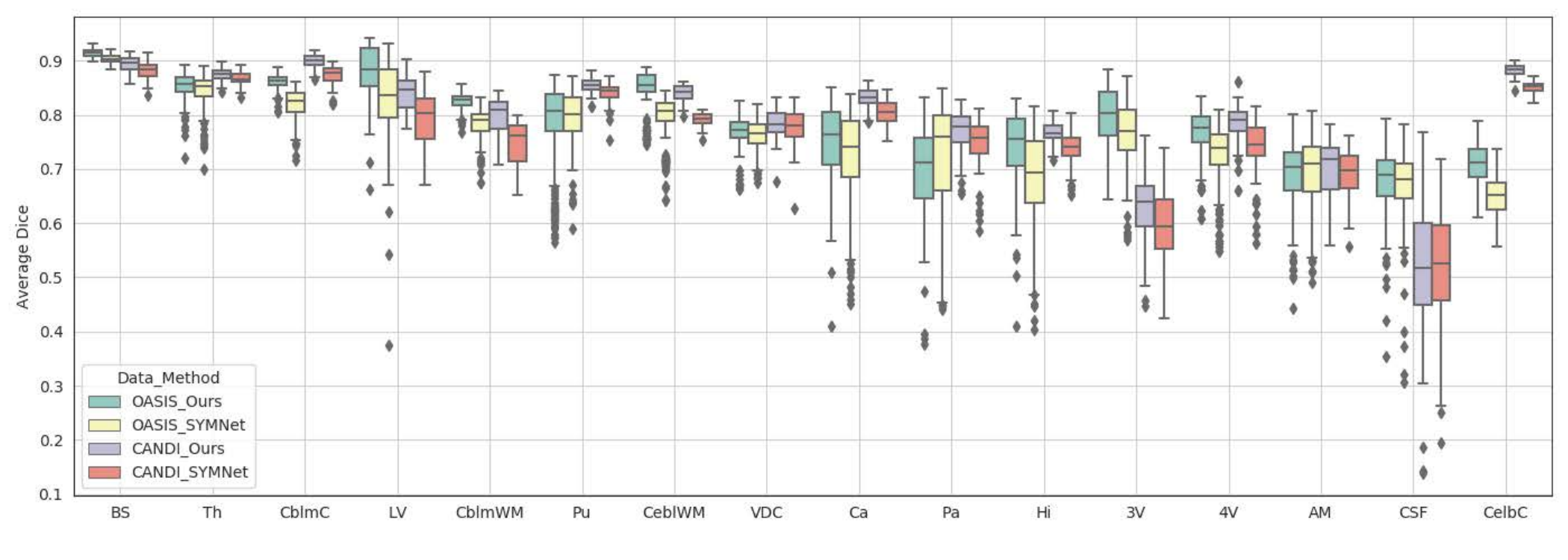}
	\end{center}
	\caption{
		Boxplots indicating Dice for 28 anatomical structures on OASIS and CANDI datasets for SYMNet and our method. 
		The abbreviations here represent brain stem (BS), thalamus (Th), cerebellum cortex (CblmC), lateral ventricle (LV), cerebellum white matter (CblmWM), putamen (Pu), cerebral white matter (CeblWM), Ventral DC (VDC), caudate (Ca), pallidum (Pa), hippocampus (Hi), 3rd ventricle (3V), 4rd ventricle (4V), amygdala (AM), CSF (CSD), cerebral cortex (CelbC).}
	\label{fig:SOTA_errorbar}
\end{figure*}

The details of 28 anatomical structures on which dice scores are calculated are provided in Figure~\ref{fig:SOTA_errorbar}. For both the OASIS and CANDI datasets, our method demonstrates a consistent and significant improvement in both mean dice scores over the all brain structures and on each anatomical category as shown in Figure~\ref{fig:SOTA_errorbar}.

\subsection{Qualitative comparison}
\begin{figure*}[htb]	
    \small
	\begin{center}
		\includegraphics[width=0.95\linewidth,keepaspectratio]{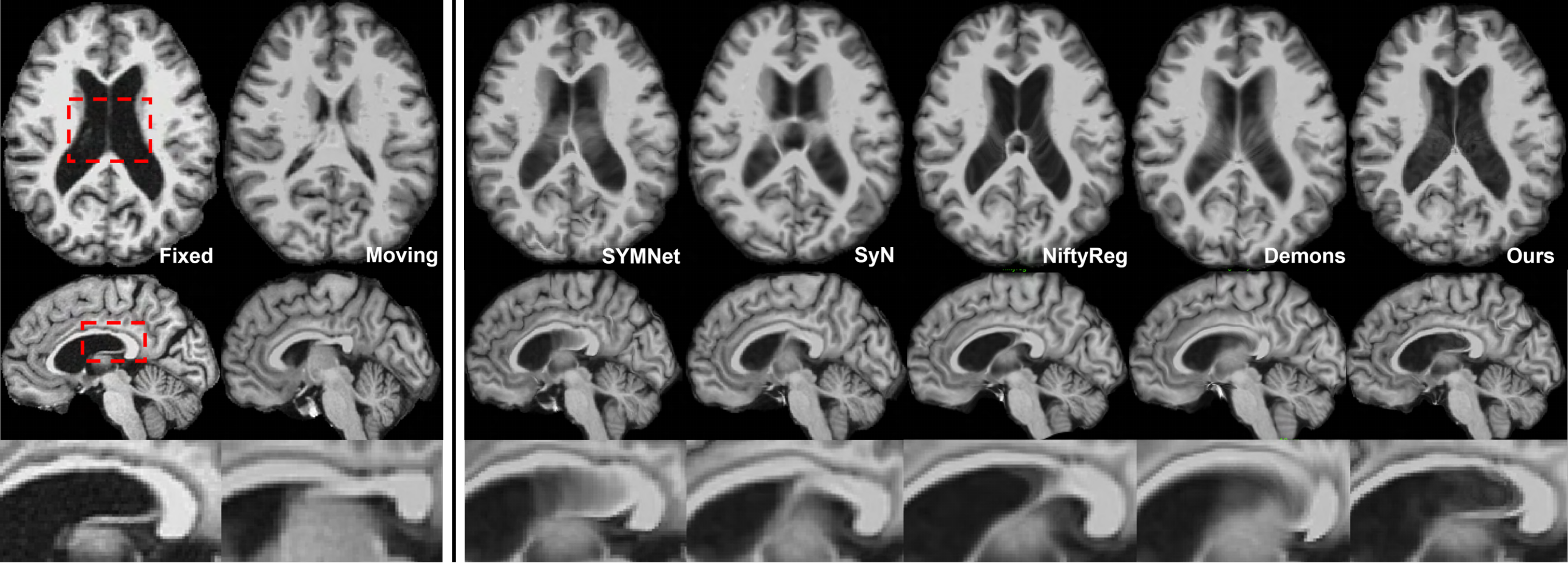}
	\end{center}
	\caption{
		Images showing an example of a registration image pair. Fixed image is OASIS ID001 and Moving image is OASIS ID002. The 3rd column to the 7th column are results of SYMNet, SyN, NiftyReg, Log-Demons and ours respectively.}
	\label{fig:full}
\end{figure*}

We present the qualitative results of SYMNet and ours in Figure~\ref{fig:img_eg}. In this supplementary material, we provide qualitative comparisons of full benchmarks in Figure~\ref{fig:full}.

\clearpage

\end{document}